%% file: root.tex
\documentclass[letterpaper, 10 pt, conference]{ieeeconf}  

\IEEEoverridecommandlockouts                              

\usepackage{graphics} 
\usepackage{epsfig} 
\usepackage{subfig}
\usepackage{mathptmx} 
\usepackage{times} 
\usepackage{amsmath} 
\usepackage{amssymb}  
\let\proof\relax
\let\endproof\relax
\usepackage{amsthm}
\usepackage{newtxtext,newtxmath}
\usepackage{optidef}
\usepackage{algorithm}
\usepackage[noend]{algpseudocode}
\usepackage{xcolor}
\usepackage{empheq}
\usepackage{flushend}
\usepackage{comment}
\usepackage{soul}

\title{\LARGE \bf
Real-Time-Feasible Collision-Free Motion Planning For Ellipsoidal Objects
}

\author{Yunfan Gao$^{1, 2}$, Florian Messerer$^{2}$, Niels van Duijkeren$^1$, Boris Houska$^{3}$, Moritz Diehl$^{2,4}$
\thanks{$^{1}$ Robert Bosch GmbH, Corporate Research, Stuttgart, Germany
	{\tt\small \{yunfan.gao, niels.vanduijkeren\}@de.bosch.com}}
\thanks{$^{2}$ Department of Microsystems Engineering (IMTEK), University of Freiburg, Germany
	{\tt\small \{florian.messerer, moritz.diehl\}@imtek.uni-freiburg.de}}
\thanks{$^{3}$ School of Information
Science and Technology, ShanghaiTech University, Shanghai, China {\tt\small borish@shanghaitech.edu.cn} }
\thanks{$^{4}$ Department of Mathematics, University of Freiburg, Germany}
\thanks{The research that led to this paper was funded by Robert Bosch GmbH.
This work was also supported by DFG via Research Unit FOR 2401, project 424107692 and 525018088, by BMWK via 03EI4057A and 03EN3054B, and by the EU via ELO-X 953348.}
}

\theoremstyle{definition}
\newtheorem{definition}{Definition}
\newtheorem{theorem}{Theorem}
\newtheorem{corollary}[theorem]{Corollary}
\newtheorem{lemma}[theorem]{Lemma}
\newtheorem{remark}[theorem]{Remark}

\DeclareMathOperator{\intSet}{int}

\DeclareMathSymbol{\shortminus}{\mathbin}{AMSa}{"39}

\newcommand{\ellipsoidE}[2]{\mathcal{E}\left({#1}, {#2}\right)}
\newcommand{\bbR}{\mathbb{R}}

\newcommand{\obsM}{M}
\newcommand{\robotM}{G}

\newcommand{\Compactcdots}{\mathinner{\cdotp\mkern-2mu\cdotp\mkern-2mu\cdotp}}

\begin{document}

\IEEEaftertitletext{\vspace{-5pt}}
\maketitle
\thispagestyle{empty}
\pagestyle{empty}

\begin{abstract}

Online planning of collision-free trajectories is a fundamental task for robotics and self-driving car applications.
This paper revisits collision avoidance between ellipsoidal objects
    using differentiable constraints.
Two ellipsoids do not overlap if and only if the endpoint of the vector between
    the center points of the ellipsoids does not lie in the interior of the Minkowski sum of the ellipsoids.
This condition is formulated using a
    parametric over-approximation of the Minkowski sum,
    which can be made tight in any given direction.
The resulting collision avoidance constraint is included in an optimal control problem (OCP)
    and evaluated in comparison to the separating-hyperplane approach.
Not only do we observe that the Minkowski-sum formulation is computationally more efficient in our experiments, but also that
    using pre-determined over-approximation parameters based on warm-start trajectories leads to a very limited increase in suboptimality.
This gives rise to a novel real-time scheme for collision-free motion planning with model predictive control~(MPC).
Both the real-time feasibility and the effectiveness of the constraint formulation are demonstrated in challenging real-world experiments.

\end{abstract}


\setlength{\abovedisplayskip}{0.5ex plus0pt minus1pt}
\setlength{\belowdisplayskip}{0.5ex plus0pt minus1pt}
\setlength{\belowcaptionskip}{-5pt}

\input{sections/1-introduction.tex}
\input{sections/2-probStatement.tex}
\input{sections/3-formulation.tex}
\input{sections/4-sqp.tex}
\input{sections/5-results.tex}

\section{CONCLUSIONS}

This paper presented a constraint formulation for collision-free motion planning for ellipsoidal objects,
    achieving non-conservative collision avoidance through a parametric over-approximation of the Minkowski sum of ellipsoids.
Updating the over-approximation parameters online and fixing their values in the OCPs leads to significant computation time savings
while retaining the capability of navigation through narrow passages.
The effectiveness of the constraint formulation is demonstrated in simulation and on an actual differentiable-drive robot.
Future work might aim at analyzing the degree of suboptimality caused by the fixed over-approximations and evaluating the method for dynamic obstacles.





\bibliographystyle{ieeetr}

\bibliography{refs.bib} 

\end{document}

%% file: sections/1-introduction.tex
\section{INTRODUCTION}
\vspace{-2pt}
Collision-free motion planning for robotic systems receives significant attention in real-world applications.
Solving optimal control problems (OCPs) sufficiently fast enables a model predictive control (MPC) scheme to promptly replan online and react to the environment in a timely manner.
The constraint formulation in OCPs, in particular collision avoidance constraints, greatly impacts computational efficiency.

A substantial body of work has investigated differentiable collision avoidance constraints for non-circular objects.
The authors of~\cite{Brito2019} address the collision avoidance between a circle and an ellipsoid
    by offline predetermining an ellipsoid that covers the collision region with minimally enlarged semi-axes.
The enlarged ellipsoid, however, does not provide a tight bound of the collision region in every direction.
With respect to polytopic objects, the authors of~\cite{Zhang2021} use duality theory to formulate collision-avoidance constraints.
The impact of shape representation on computational efficiency is investigated in~\cite{Dietz2023}.
The authors conclude that a vertex-based representation of polyhedra results in shorter solution times than a half-space-based representation.
Additionally, in~\cite{Reiter2024}, the authors smoothen rectangle obstacles progressively by $p$-norms over the OCP prediction horizon.

In this paper, we focus on collision avoidance between ellipsoidal objects.
Ellipsoids are a helpful geometric object in collision-free motion planning, not only to approximate robot shapes,
    but also to model uncertainties for point objects~\cite{Gao2023}.
Ellipsoidal calculus has been investigated in many different areas,
    e.g., ellipsoid packaging~\cite{Birgin2015} and collision detection in computer graphics~\cite{Schneider2004}.
The distance computation between non-overlapping ellipsoids is a convex optimization problem.
In the 2D case, the distance can be computed by solving a polynomial equation~\cite{Eberly2008}.
The optimization problem of signed-distance computation for overlapping ellipsoids is non-convex.
The authors of~\cite{Iwata2015} propose to compute the signed distance
   by identifying the points that satisfy relaxed Karush-Kuhn-Tucker (KKT) conditions and subsequently evaluating the signed distance values at these points.

To detect the overlapping of ellipsoids,
    the authors of~\cite{Iwata2015, Alfano2003}, and~\cite[Section 11.9.2]{Schneider2004} shrink or grow the two ellipsoids until they share exactly one point.
The scale factor being greater than one indicates that the intersection is an empty set.
Although this condition is useful in overlapping detection,
    it is not differentiable.
Differentiable collision avoidance constraints are formulated in~\cite{Birgin2015, Kallrath2013, Nocedal2006}
    by ensuring the existence of a separating hyperplane~\cite[Theorem 11.3]{Rockafellar1970}.
To simplify the constraint formulation,
    the authors of~\cite{Birgin2015} transform one of the two ellipsoids into a unit circle (or sphere) via an affine transformation.
The collision-free condition can be ensured by imposing that the distance from the circle center to the scaled ellipsoid is no smaller than one.

In the present paper, we ensure collision avoidance by using a different necessary and sufficient condition for two ellipsoids not to overlap.
The condition requires the endpoint of the vector between the center of the two ellipsoids to be located outside or on the boundary of their Minkowski sum centered at the origin.
We numerically formulate this condition via a parametric over-approximation of the Minkowski sum,
    which can be made tight in any direction.
By leaving the parameter of the over-approximation as an optimization variable,
    we achieve collision avoidance without introducing extra separation distance.
For the resulting formulation, we observe a better computational performance compared to the separating-hyperplane approach.

Furthermore, we facilitate real-time capability by computing the over-approximation parameters based on the latest OCP solution and keeping their values fixed while solving the OCP.
This gives rise to suboptimality,
    but values for the fixed parameters resulting in minor suboptimality can be obtained with little computational effort.
Thereby, the computational complexity is reduced without compromising the capability of the robot navigating through cluttered environments.

The OCP formulation is evaluated in a model predictive control (MPC) scheme both in simulation and on a physical differential-drive robot.
We show that planning collision-free trajectories using the proposed constraint formulation is real-time feasible.
The main contributions of this work are:
\begin{enumerate}
    \item A computationally efficient formulation of differentiable collision-avoidance constraints for ellipsoidal objects.
    Non-conservative collision avoidance can be achieved despite the over-approximation of the Minkowski sum of ellipsoids.
    \item Improvement of real-time feasibility at an insignificant cost of robot motion inefficiency by updating the over-approximation parameters outside the OCP.
    \item Experimental validation in an environment cluttered with virtual ellipsoidal obstacles.
\end{enumerate}

\subsubsection*{Notation}
In this paper, when referring to collision avoidance, we allow that two sets \textit{touch}, but the interior of the two sets must not intersect.
The interior of a set $\mathcal{B}$ is denoted by $\intSet\mathcal{B}$.
The sequence of natural numbers for an interval $\left[a, b \right]$ is denoted by $\mathbb{N}_{\left[a, b \right]}$.
An ellipsoid in $\bbR^n$ is a set of the form
\begin{equation}
    \ellipsoidE{t}{M} := \left\lbrace \tau \in \bbR^n \vert (\tau-t) M^{-1} (\tau - t) \leq 1 \right\rbrace,
\end{equation}
where $t \in \bbR^n $ is the center of the ellipsoid and $M \in\mathbb{S}^{n}_{++}$ is a positive-definite matrix.

%% file: sections/2-probStatement.tex
\section{MOTION PLANNING PROBLEM STATEMENT}
\label{sec:prob-statement}

In the following, we describe the problem set-up and define the OCP we want to solve.
\vspace{-5pt}
\subsection{Robot System Dynamics}
\vspace{-2pt}
Let $x\in\bbR^{n_x}$ and $u\in\bbR^{n_u}$ be the robot system state and the control input respectively.
The discrete (or discretized) system dynamics take the form
\begin{equation}
	x_{k+1} = \psi_k(x_k, u_k),\; k \in \mathbb{N}_{[0, N-1]}.
\end{equation}
Due to the physical limitations of the system and control design objectives such as recursive feasibility,
    the trajectories are subject to state-input constraints ${g_k(x_k, u_k) \leq 0}$ and terminal constraints ${g_N(x_N) \leq 0}$.
The functions $\psi_k$, $g_k$, and $g_N$ are twice continuously differentiable in all arguments.

\vspace{-5pt}
\subsection{Robot Shape Modeling}
\vspace{-2pt}
The robot shape is modeled by an ellipsoid $\ellipsoidE{0}{\robotM}$.
The system state $x_k$ contains information on the center position $p_{\mathrm{c}}:\bbR^{n_x}\!\to\!\bbR^{n_{\mathrm{w}}}$,
    where ${n_{\mathrm{w}} \in \left\lbrace 2, 3 \right\rbrace}$ is
the dimension of the physical world.
The mapping from the robot state to the rotation matrix is denoted by ${R:\bbR^{n_x}\!\to\!\bbR^{n_{\mathrm{w}} \times n_{\mathrm{w}}}}$.
Given the robot state $x_k$, the space occupied by the robot is a rotation and translation of the ellipsoid $\ellipsoidE{0}{\robotM}$, which is given by $\ellipsoidE{p_{\mathrm{c}}(x_k)}{R(x_k)\robotM R(x_k)^\top}$.
To simplify notation, let $\tilde{\robotM}(x_k) := R(x_k)\robotM R(x_k)^\top$.

\vspace{-5pt}
\subsection{Obstacle Avoidance}
\vspace{-2pt}
Consider a set of ellipsoidal obstacles:
$ \lbrace \ellipsoidE{t_1}{\obsM_1}, \allowbreak  \ellipsoidE{t_2}{\obsM_2},  \cdots, \ellipsoidE{t_{n_m}}{\obsM_{n_m}}\rbrace$.
We aim to ensure that for each time index ${k \!\in\! \mathbb{N}_{[0, N]}}$ and each obstacle $m \!\in\! \mathbb{N}_{[1, n_m]}$,
   the interior of the robot and the interior of the obstacle do not intersect.
Let ${l}_k\left(x_k, u_k\right)$ and ${l}_N\left(x_k\right)$ be the stage cost and terminal cost functions,
    which are twice continuously differentiable.
Let $\bar{x}_0$ be the current robot state.
The discrete-time optimal control problem (OCP) is formulated as follows:
\vspace{-5pt}
\begin{mini!}|s|
	{\substack{x_0, \Compactcdots, x_N,\\ u_0, \Compactcdots, u_{N\!\shortminus \!1}}}
	{\sum_{k=0}^{N\!-\!1} {l}_k\left(x_k, u_k\right) + {l}_N\left(x_N\right) \label{eq:ocp-objective-function}}
	{\label{eq:ocp-prob-statement}}{}
	\addConstraint{x_0}{=\bar{x}_0}
	\addConstraint{x_{k+1}}{=\psi_k(x_k, u_k),\, k\in \mathbb{N}_{[0, N \shortminus 1]}}
    \addConstraint{0}{\geq g_k(x_k, u_k),\, k\in \mathbb{N}_{[0, N \shortminus 1]}}
    \addConstraint{0}{\geq g_N(x_N)}
    \addConstraint{\varnothing}{ = \intSet\left(  \ellipsoidE{p_{\mathrm{c}}(x_k)}{\tilde{\robotM}(x_k)} \right) \cap  \intSet\left( \ellipsoidE{t_{m}}{\obsM_{m}} \right) \nonumber}
    \addConstraint{}{\quad k\in \mathbb{N}_{[0, N]}, m\in\mathbb{N}_{[1, n_m]} \label{eq:constr-intersection-empty} }.
\end{mini!}

%% file: sections/3-formulation.tex
\section{COLLISION AVOIDANCE CONSTRAINT FORMULATION}
\label{sec:formulation}

In this section, we first present the preliminary knowledge related to the proposed constraint formulation and then derive the Minkowski-sum-based constraint formulation.

\vspace{-2pt}
\subsection{Preliminaries}
\vspace{-1pt}
\label{sec:preliminary}
Supporting functions are an important tool for the analysis of convex sets.
Minkowski sums of ellipsoids have been analyzed in many contexts, in particular reachable sets for dynamic systems and robust optimization~\cite{Houska2011}.
Here we present the preliminary knowledge of the supporting function and the Minkowski sum related to our constraint formulation.

\begin{definition}[Supporting function and supporting halfspace]
    Given a closed convex set $\mathcal{B} \subset \mathbb{R}^n$,
        the supporting function $V(\eta; \mathcal{B})$ and the supporting halfspace $\mathcal{H}(\eta; \mathcal{B})$ for a direction $\eta \in\bbR^{n} \setminus \lbrace 0 \rbrace$ are given by
    \begin{subequations}
    \begin{align}
        V(\eta; \mathcal{B}) &:= \max_{b \in \mathcal{B}} \eta^\top b ,\\
        \mathcal{H}(\eta; \mathcal{B}) &:= \left\lbrace b\in\bbR^{n} \vert \eta^\top b \leq V(\eta; \mathcal{B}) \right\rbrace .
    \end{align}
    \end{subequations}
\end{definition}

For an ellipsoid $\ellipsoidE{0}{M}$, the supporting function can be expressed in closed form~\cite[Example 2.10]{Houska2011}:
\begin{equation}
    \label{eq:ellip-supporting-func}
    V(\eta; \ellipsoidE{0}{M}) = \sqrt{\eta^{\top} M \eta}.
\end{equation}

\begin{definition}[Minkowski sum]
    The Minkowski sum of two sets $\mathcal{B} \subset \mathbb{R}^n$
    and $\mathcal{D}\subset \mathbb{R}^n$ is defined as
    \begin{equation}
        \mathcal{B} \oplus \mathcal{D}:=
    \left\lbrace b+d \, \vert \, b \in \mathcal{B} \; \mathrm{and} \; d \in \mathcal{D} \right\rbrace.
    \label{eq:minkowskiSum-def}
    \end{equation}
\end{definition}

\begin{lemma}
    \label{lemma:ellipsoid-over-approximation}
    The Minkowski sum of the two ellipsoids can be over-approximated by a third ellipsoid:
\begin{equation}
    \ellipsoidE{0}{M_1} \oplus \ellipsoidE{0}{M_2} \subseteq \ellipsoidE{0}{\frac{M_1}{\beta_1}+\frac{M_2}{\beta_2}},
    \label{eq:ellipsoid-over-approximation}
\end{equation}
for any $\beta_1, \beta_2 > 0$ satisfying $\beta_1 + \beta_2 = 1$.
\end{lemma}

\proof
The main idea of the proof is that for any direction $\eta \in {\bbR^n \setminus \lbrace 0 \rbrace}$,
    the value of the supporting function associated with the right-hand side of~\eqref{eq:ellipsoid-over-approximation} is greater than or equal to its counterpart of the left-hand side.
Therefore, the supporting halfspace of the left-hand side is a subset of its counterpart of the right-hand side.
Since a closed convex set can be represented by the intersection of its supporting halfspaces,
    Lemma~\ref{lemma:ellipsoid-over-approximation} can be derived
    (see the detailed proof in~\cite[Theorem 2.4]{Houska2011}).
\endproof
\vspace{-5pt}

\begin{lemma}
\label{lemma:over-approximation-tight}
Given one direction $\eta \in {\bbR^n \setminus \lbrace 0 \rbrace}$,
    there exists $\beta^*_1, \beta^*_2 > 0$ satisfying $\beta^*_1 + \beta^*_2 = 1$ such that the over-approximation is tight in the direction of $\eta$:
\begin{equation}
    V\left(\eta; \ellipsoidE{0}{\frac{M_1}{\beta_1^*}+\frac{M_2}{\beta_2^*}}\right) = V\left(\eta; \ellipsoidE{0}{M_1} \oplus \ellipsoidE{0}{M_2}\right).
\end{equation}
\end{lemma}
\proof
We refer to~\cite[Theorem 4.1]{Kurzhanskiy2007} for the proof.
\endproof
\vspace{-5pt}

\begin{figure}[t]
	\centering
	\subfloat[Minkowski sum (colored in green) of two ellipsoids (colored in blue) and different over-approximations (colored in orange).
    The solid black line depicts the boundary of the supporting halfspace of the Minkowski-sum,
        which coincides with the over-approximation counterpart for ${\gamma=0.24}$.
    The intersections between the ellipsoid $\mathcal{E}(t_2, M_2)$ and the Minkowski sum (and its over-approximations) are irrelevant.]
	{\includegraphics[width=0.98\linewidth]
		{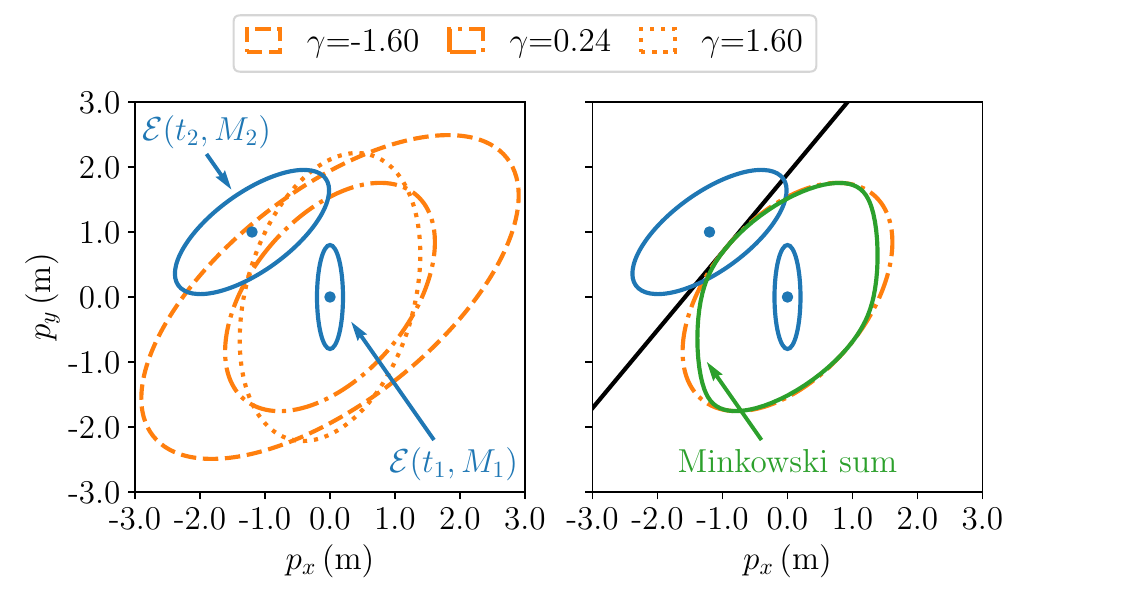}
		\label{fig:tightness-space}}\\
	\vspace{-6pt}
	\subfloat[The over-approximation for ${\gamma=0.24}$ achieves the minimum value of the supporting function (the same value as the Minkowski sum counterpart).]
	{\includegraphics[width=0.98\linewidth]
		{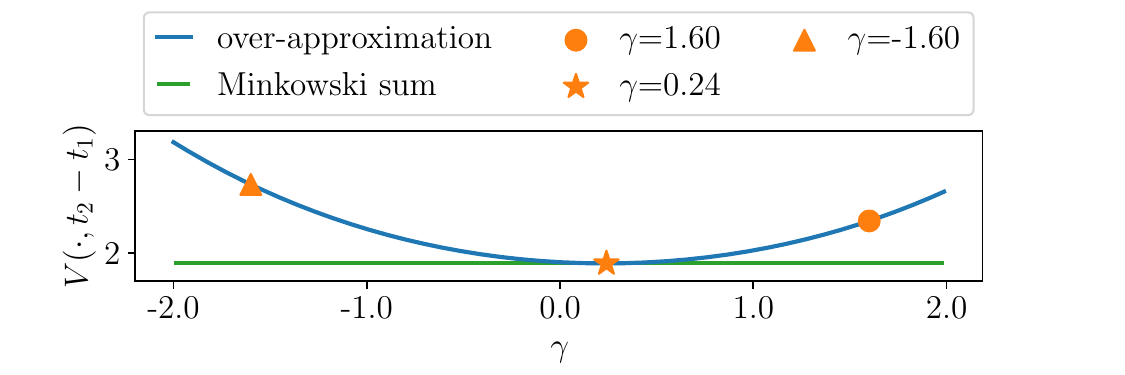}
		\label{fig:tightness-versus-beta}}
	\caption{Illustrative example}
	\label{fig:illustrative-example}
	\vspace{-12pt}
\end{figure}

An illustrative example demonstrating the tightness of over-approximations is shown in Fig.~\ref{fig:illustrative-example}.

\vspace{-2pt}
\subsection{Minkowski-sum-based Constraint Formulation}
\vspace{-2pt}
\begin{lemma}
    \label{lemma:equal-symmetric}
    Consider two ellipsoidal sets $\ellipsoidE{t_1}{M_1}$ and $\ellipsoidE{t_2}{M_2}$.
    The interior of the two sets does not intersect if and only if the point $t_1 \shortminus t_2$ is not in the interior of the Minkowski sum of two ellipsoids centered at the origin:
    \begin{equation}
        \begin{split}
             \varnothing &= \intSet\left( \ellipsoidE{t_1}{M_1}\right) \cap \intSet\left( \ellipsoidE{t_2}{M_2} \right) \Leftrightarrow\\
            t_1 \shortminus t_2 &  \notin \intSet\left( \ellipsoidE{0}{M_1} \oplus \ellipsoidE{0}{M_2} \right).
        \end{split}
    \end{equation}
\end{lemma}

\proof
Lemma~\ref{lemma:equal-symmetric} can be derived from the observation that an ellipsoid is symmetric with respect to its center:
{
\allowdisplaybreaks
\begin{subequations}
    \begin{align}
        &  t_1 \shortminus t_2 \in \intSet\left( \ellipsoidE{0}{M_1} \oplus \ellipsoidE{0}{M_2} \right), \\
        \Leftrightarrow\; & \text{there exists}\; \tau_1, \tau_2 \in \bbR^{n} \; \text{such that}\; \tau_1+\tau_2 = t_1 \shortminus t_2, \nonumber\\
        & \quad\tau_1^\top M_1^{\shortminus 1}\tau_1< 1, \;\text{and}\; \tau_2^\top M_2^{\shortminus 1}\tau_2< 1, \\
        \Leftrightarrow\; & \text{there exists}\; \tau_1, \tau_2^{\prime} \in \bbR^{n} \; \text{such that}\; \tau_1 \shortminus  \tau_2^{\prime} = t_1 \shortminus t_2, \nonumber\\
        & \quad\tau_1^\top M_1^{\shortminus 1}\tau_1< 1, \;\text{and}\; \left(\shortminus \tau_2^{\prime}\right)^{\top} M_2^{\shortminus 1}\left(\shortminus \tau_2^{\prime}\right)< 1, \\
        \Leftrightarrow\; & \text{there exists}\; \tau_1 \in \bbR^{n} \; \text{such that}\; \tau_1^\top M_1^{\shortminus 1}\tau_1< 1 \; \text{and}  \nonumber\\
        & \quad\left(t_1 \shortminus t_2 \shortminus \tau_1\right)^{\top} M_2^{\shortminus 1}\left(t_1 \shortminus t_2 \shortminus \tau_1\right)< 1,  \\
        \Leftrightarrow\; & \varnothing \neq \intSet\left( \ellipsoidE{0}{M_1}\right) \cap \intSet\left( \ellipsoidE{t_1 \shortminus t_2}{M_2} \right), \\
        \Leftrightarrow\; & \varnothing \neq \intSet\left( \ellipsoidE{t_1}{M_1}\right) \cap \intSet\left( \ellipsoidE{t_2}{M_2} \right).
    \end{align}
\end{subequations}
}
\endproof

\begin{corollary}
\label{thm:over-approx-same-dist}
Given two ellipsoidal sets $\ellipsoidE{t_1}{M_1}$ and $\ellipsoidE{t_2}{M_2}$,
the interior of the two ellipsoidal sets does not intersect
if and only if there exists at least one over-approximation such that the point $t_1 \shortminus t_2$ is not in the interior of the over-approximation.
\begin{equation}
    \begin{split}
        \varnothing\! =\! & \intSet\left( \ellipsoidE{t_1}{M_1}\right) \! \cap  \intSet\left( \ellipsoidE{t_2}{M_2} \right) \! \Leftrightarrow \exists\; \beta_1^*, \beta_2^* > \!0\colon \\
        & \beta_1^* + \beta_2^* \! = \! 1, \, t_1 \shortminus t_2 \notin \intSet\left( \ellipsoidE{0}{\frac{M_1}{\beta_1^*}+\frac{M_2}{\beta_2^*}} \right).
    \end{split}
\end{equation}
\end{corollary}
\proof
Corollary~\ref{thm:over-approx-same-dist} follows from Lemma~\ref{lemma:over-approximation-tight} and Lemma~\ref{lemma:equal-symmetric}.
\endproof

For computational considerations,
    we substitute the variables $\beta_1$ and $\beta_2$ by introducing a new variable $\gamma\in\bbR$:
\begin{equation}\label{eq:gamma-beta-relation}
    \beta_1 = \frac{1}{1 + \exp(\gamma)} \; \text{and} \;
    \beta_2 = \frac{1}{1 + \exp(\shortminus \gamma)}.
\end{equation}
The constraints on $\beta_1$ and $\beta_2$ are satisfied for any $\gamma\in\bbR$:
\begin{equation*}
    \begin{split}
        &\frac{1}{1 + \exp(\gamma)} > 0,\; \frac{1}{1 + \exp(\shortminus \gamma)} > 0, \\
        &\frac{1}{1 + \exp(\gamma)} +  \frac{1}{1 + \exp(\shortminus \gamma)}
        =  \frac{1}{1 + \exp(\gamma)} +  \frac{ \exp(\gamma)}{1 + \exp(\gamma)} = 1.
    \end{split}
\end{equation*}
Constraint~\eqref{eq:constr-intersection-empty} is thereby reformulated
    using an additional optimization variable $\gamma_{m, k}$ for each obstacle $m$ and every time index $k$ as follows:
\begin{equation}
\label{eq:constr-formulation}
    \begin{split}
        1 \leq & \left(p_{\mathrm{c}}(x_k) \! \shortminus \! t_m\right)^{\!\top} \left(\left(1+\exp(\gamma_{m, k})\right)\tilde{G}(x_k) \right. \\
        & \quad +\left. \left(1+\exp(\shortminus\gamma_{m, k})\right) \obsM_m \right)^{\shortminus 1} \left(p_{\mathrm{c}}(x_k) \! \shortminus \! t_m\right).
    \end{split}
\end{equation}

\begin{remark}
    This formulation can be extended to zonotopes.
    A zonotope can be viewed as the Minkowski sum of a set of line segments,
        i.e., degenerate ellipsoids whose corresponding $M$ matrices are positive \textit{semi}-definite.
    The constraints for zonotope objects can thereby be formulated by taking nested over-approximations, one on the object shape and the other on the Minkowski sum of the two objects.
\end{remark}

\subsection{Bounds on Optimization Variable $\gamma$}
\label{sec:gamma-bound}
Numerical solvers may encounter difficulties
    when the values of $\exp(\gamma)$ and $\exp(\shortminus\gamma)$ are close to zero or very large.
A bound on the variable $\gamma$ can be imposed to improve numerical robustness.
In this context, we derive an appropriate bound for $\gamma$ such that the optimality is unaffected.

Recall the closed-form expression for the value of the supporting function with respect to an ellipsoid~\eqref{eq:ellip-supporting-func}.
For an over-approximation given by $\left(1+\exp(\gamma)\right)M_1+ \left(1+\exp(\shortminus\gamma)\right)M_2$,
    the value of $\gamma^*$ that obtains the minimum value of the supporting function is given by:
\begin{equation*}
    \gamma^*(\eta) := \arg\min \sqrt{\left(1 \!+ \exp(\gamma) \right) \eta^\top M_1 \eta + \left(1 \!+ \exp(\shortminus \gamma) \right)\eta^\top M_2 \eta}.
\end{equation*}
The supporting function is a convex function of $\gamma$ given a direction $\eta$.
The optimal $\gamma^*(\eta)$ can be expressed in closed form:
\begin{equation}
    \gamma^*(\eta) = \frac{1}{2} \log \left( \frac{\eta^\top M_2 \eta}{\eta^\top M_1 \eta} \right).
    \label{eq:closed-form-gamma}
\end{equation}
We have
$\frac{\lambda_{\min}(M_2)}{\lambda_{\max}(M_1)} \leq \frac{\eta^\top M_2 \eta}{\eta^\top M_1 \eta} \leq \frac{\lambda_{\max}(M_2)}{\lambda_{\min}(M_1)}$,
where $\lambda_{\min}$ and $\lambda_{\max}$ denote the least and the largest eigenvalues.
Therefore, we can impose a bound on $\gamma$ for numerical robustness without compromising the optimality of the over-approximation.
The bound on $\gamma$ is given by
\begin{equation}
    \left[\frac{1}{2}\log\left(\frac{\lambda_{\min}(M_2)}{\lambda_{\max}(M_1)}\right), \frac{1}{2}\log\left(\frac{\lambda_{\max}(M_2)}{\lambda_{\min}(M_1)}\right)\right].
    \label{eq:bound-gamma}
\end{equation}

%% file: sections/4-sqp.tex
\begin{figure*}[th]
	\centering
	\hfill
	\subfloat[The robot would crash into the obstacles if it stayed on the reference path.]
	{\includegraphics[width=0.3\linewidth]
		{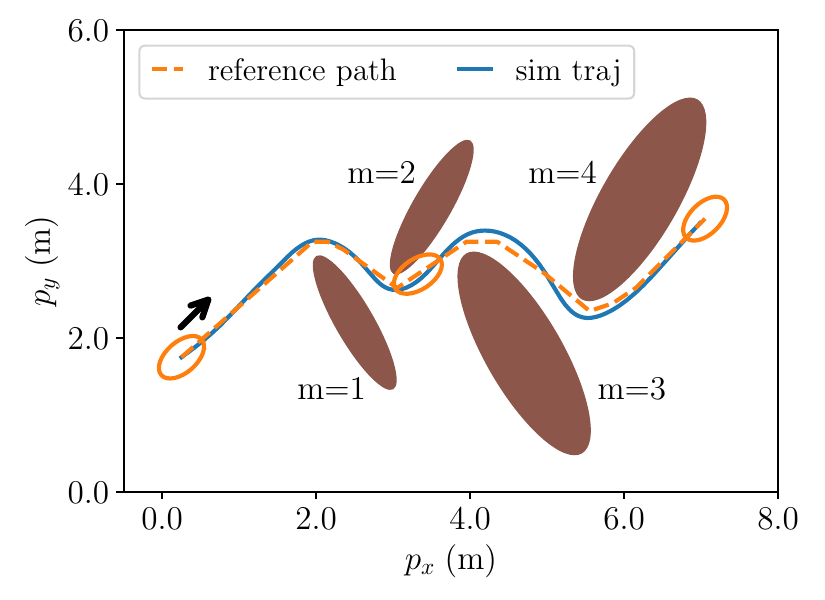}
		\label{fig:sim-environment}}
	\hfill
	\subfloat[The robot safely travels through the narrow passages.]
	{\includegraphics[width=0.3\linewidth]
		{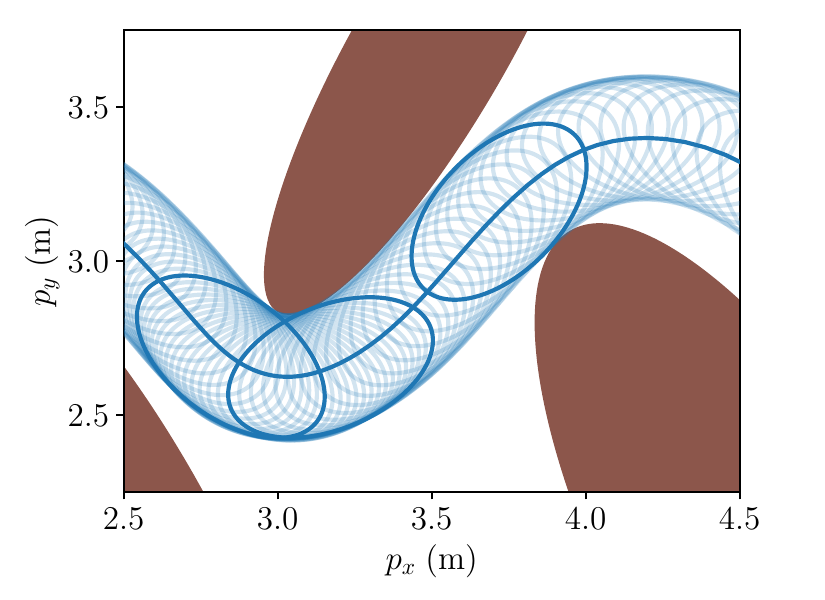}
		\label{fig:zoom-in-sim-trajectory}}
	\hfill
	\subfloat[Distance to the obstacles over time.]
	{\includegraphics[width=0.3\linewidth]
		{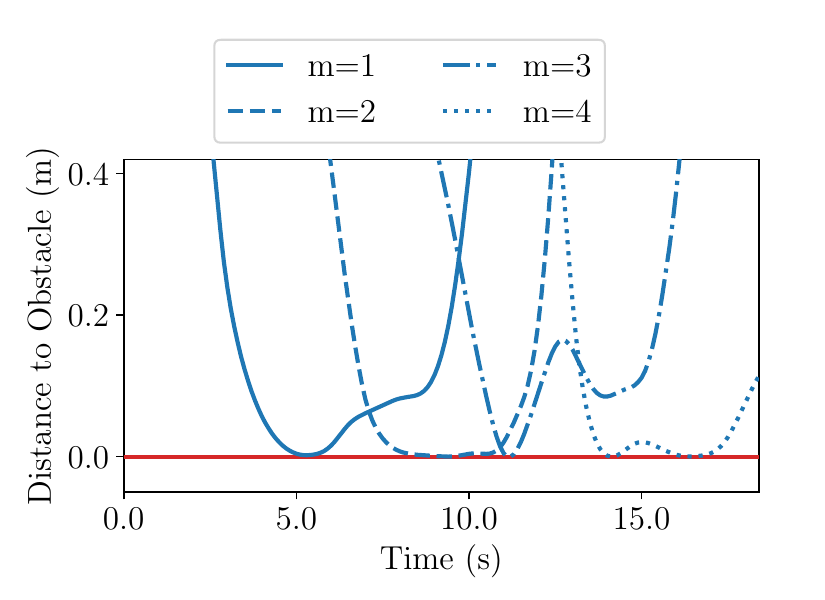}
		\label{fig:distance-over-time}}
	\hfill
	\caption{MPC simulation results.
		The over-approximations are optimized within the OCPs.
		The solid-line ellipsoids depict the differential-drive robot at different time steps, and the ellipsoids colored in brown depict the obstacles to avoid.}
	\label{fig:sim-results}
	\vspace{-5pt}
\end{figure*}

\vspace{-3pt}
\section{REAL-TIME NUMERICAL OPTIMIZATION}
\label{sec:practical-considerations}
\vspace{-1pt}
In this section, we present several approaches to improve the numerical properties and facilitate real-time applications.

\vspace{-3pt}
\subsection{Fixed Parameterization of Over-approximation}\label{sec:practical-considerations-fixed-gamma}
\vspace{-1pt}
It can be seen from Corollary~\ref{thm:over-approx-same-dist} that
    for any over-approximation of the Minkowski-sum of two ellipsoids $\ellipsoidE{t_1}{M_1}$ and $\ellipsoidE{t_2}{M_2}$, the point $t_1 \shortminus t_2$ not being in the interior of the over-approximation is a sufficient condition for the interior of the two ellipsoidal sets not to intersect.
Therefore, solving the OCP where the over-approximation, i.e., the value of $\gamma$, is fixed still yields a collision-free trajectory.
Meanwhile, fixing $\gamma$ reduces the nonlinearity of the collision-avoidance constraint~\eqref{eq:constr-formulation}.
This comes at the cost of obtaining a suboptimal solution.
The degree of suboptimality depends on the proximity of the fixed over-approximation to the optimal over-approximation.

One option to determine the fixed over-approximation is to take the one that minimizes the corresponding supporting function in the direction of the robot center (given the solution of the last time step) to the obstacle center.
Recall that for any given non-zero direction,
    a closed-form expression exists for the value of $\gamma$ achieving the minimum projection length~\eqref{eq:closed-form-gamma}.
For each obstacle $m$ and each time index $k$,
    we compute the parameter $\hat{\gamma}_{m, k}$ for the fixed over-approximation as follows:
\begin{equation}
    \hat{\gamma}_{m, k} = \left.\frac{1}{2} \log \left( \frac{\eta^\top M_m \eta}{\eta^\top \tilde{G}(x_k) \eta} \right) \right\vert _{\eta= p_{\mathrm{c}}(x_k) \shortminus  t_m }.
    \label{eq:compute-fixed-gamma}
\end{equation}
Note that the computation merely consists of several matrix-vector multiplications and one logarithm operation.
The computation effort is therefore negligible compared to solving the OCPs.

\vspace{-3pt}
\subsection{Regularization}
\vspace{-1pt}
The Gauss-Newton Hessian approximation is widely used in solving quadratic programming (QP) subproblems in the sequential quadratic programming (SQP) method~\cite[Section~10.3]{Nocedal2006}.
It is computationally efficient as it depends only on the first-order derivatives of the objective function.
The constraint-related Hessian information is disregarded.
Given that the objective function~\eqref{eq:ocp-objective-function} does not depend on the optimization variables~$\gamma$,
    the Hessian blocks associated with these variables are zero.
Overly optimistic Newton steps are avoided through regularization of these Hessian blocks.

%% file: sections/5-results.tex
\vspace{-3pt}
\section{SIMULATION AND REAL-WORLD EXPERIMENTS}
\vspace{-1pt}
\label{sec:results}

We solve the OCPs with an SQP-type solver in \texttt{acados}~\cite{Verschueren2021} and use HPIPM~\cite{Frison2020} as the QP solver.
The simulation experiments are conducted on a laptop with an Intel i7-11850H processor and 32GB of RAM.
The real-world experiments are carried out on a Neobotix MP-500 differential-drive robot.
Its onboard computer is equipped with an Intel i7-7820EQ processor and 16GB of RAM.

\subsection{System Dynamics, Cost Function, and Constraints}
Consider a differential-drive robot modeled in a two-dimensional physical space.
The robot is centered at $\left( p_x, p_y \right)$ with heading $\theta$.
Forward and angular velocities are denoted by~$v$ and $\omega$, respectively:
$$x := \left[ \begin{array}{ccccc}
    p_x & p_y & \theta & v & \omega
\end{array}\right]^\top \in \bbR^5.$$
The control input $u$ consists of forward acceleration~$a$ and angular acceleration $\alpha$, i.e.,
$u := \left[ \begin{array}{cc}
	a & \alpha
\end{array} \right]^{\top}\in \bbR^2. $
The continuous-time equations of motion are
\begin{equation}
    \dot{x} = \left[ \begin{array}{ccccc}
		v\cos(\theta) & v\sin(\theta) & \omega & a & \alpha
	\end{array}\right]^\top.
    \label{eq:robot-model}
\end{equation}
In each discretization interval, the control input is modeled as zero-order hold.
The system is discretized by numerical integration (explicit Runge-Kutta integrator of order four).
The objective of the OCP is to track given state and input reference trajectories
    $\left( x^{\mathrm{ref}}_{0}, \dots, x^{\mathrm{ref}}_{N}  \right)$ and $\left( u^{\mathrm{ref}}_{0}, \dots, u^{\mathrm{ref}}_{N \shortminus 1} \right)$.
The stage costs and terminal costs are the weighted squared reference-tracking errors.
The robot is subject to affine constraints on $v$, $\omega$, $a$, and $\alpha$ due to actuator limits.
The terminal constraint requires the robot to be stationary at the terminal state, i.e.,
$\shortminus \epsilon_v \leq v \leq \epsilon_v $ and $\shortminus \epsilon_{\omega} \leq {\omega} \leq \epsilon_{\omega}$,
    where $\epsilon_v$ and $\epsilon_{\omega}$ are small positive values.

\subsection{Simulation Experiments}
\label{sec:simulation-results}

\subsubsection*{Experiment Setting}
The Theta* algorithm~\cite{Daniel2010} is employed to determine a path from a given initial position to the goal position.
The generated reference path is collision-free for point objects, but not for an ellipsoidal robot (see Fig.~\ref{fig:sim-environment}).
At every time instant of the MPC simulation, we segment one part of the path based on the current robot position.
The segmented path is then fitted by a spline.
A time-optimal reference trajectory, which takes into account the system limitations of the robot, is subsequently generated~\cite{Bobrow1985}.
The OCP parameters are collected in Table~\ref{table:param-simulation}.

At the first time step of the MPC simulation,
    the state variables are initialized by the reference trajectories.
The initial guess of the control variables, as well as the additional optimization variables introduced by the collision avoidance constraints, namely, $\gamma_{m, k}$,
    are set to zeros.
In subsequent time steps, the optimization process is warm-started using the solution from the previous time step.
\begin{table}[t]
	\centering
	\captionsetup{width=\linewidth}
	\vspace{4ex}
	\caption{OCP parameters}
	\label{table:param-simulation}
	\begin{tabular}{@{} l  r  r r @{}}
		\hline
		\hline
		& & & \\[\dimexpr-\normalbaselineskip+2pt]
		Name              & Unit         & Symbol     & Value  \\
		\hline
		& & & \\[\dimexpr-\normalbaselineskip+2pt]
		Prediction horizon &s           & $T$     & 2.0    \\
		Discretization intervals  & -  & $N$     & 20    \\
		Robot axis length & m  &       & $(0.4, 0.7)$ \\
        Number of obstacles & - & $n_m$ & 4 \\
		\hline
		\hline
	\end{tabular}
    \vspace{-3ex}
\end{table}

\begin{figure*}[th]
	\centering
    \hfill
    \begin{minipage}[b]{0.31\linewidth}
    \centering
	{\includegraphics[width=0.98\linewidth]{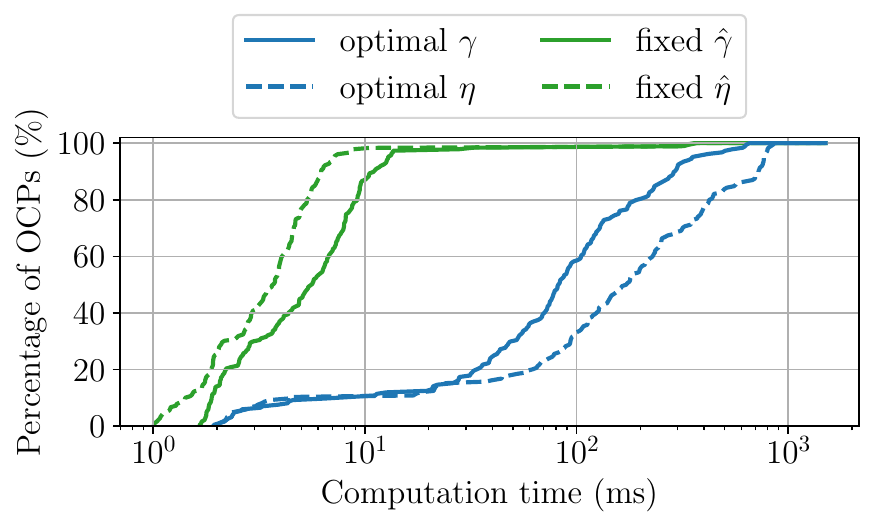}}
    \caption{Computation time for solving OCPs in MPC simulation.
    The time for computing the value of $\hat{\gamma}$ and $\hat{\eta}$ is excluded.
    The maximum number of iterations is sufficiently big for the SQP to converge.}
    \label{fig:computation-time}
    \end{minipage}
    \hfill
    \begin{minipage}[b]{0.31\linewidth}
    \centering
    \includegraphics[width=0.98\linewidth]{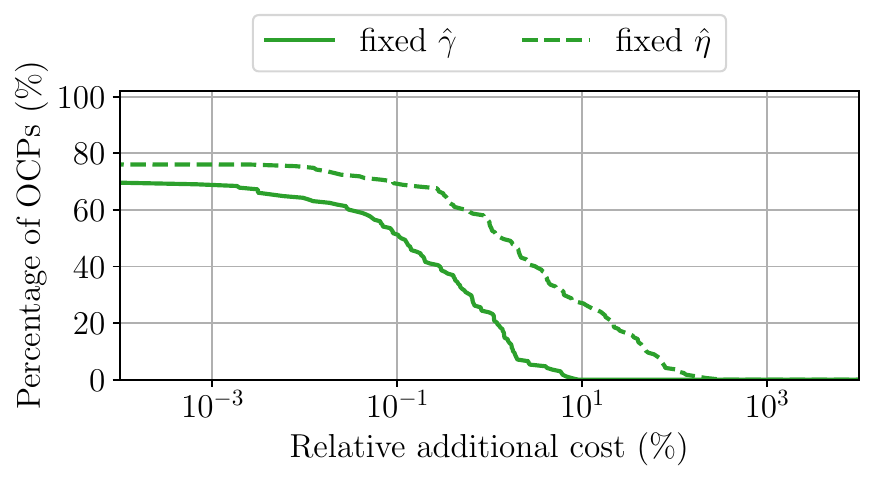}
    \caption{Relative additional cost due to fixing the over-approximation parameters $\hat{\gamma}$ and fixing the separating hyperplane parameters $\hat{\eta}$.
    The lines show the percentage of OCPs whose relative additional cost exceeds certain values.
    }
    \label{fig:fixed-gamma-eta-suboptimality}
    \end{minipage}
    \hfill
    \begin{minipage}[b]{0.31\linewidth}
    \centering
    \subfloat
	{\includegraphics[trim={0 1.2cm 0 0},clip, width=0.98\linewidth]
		{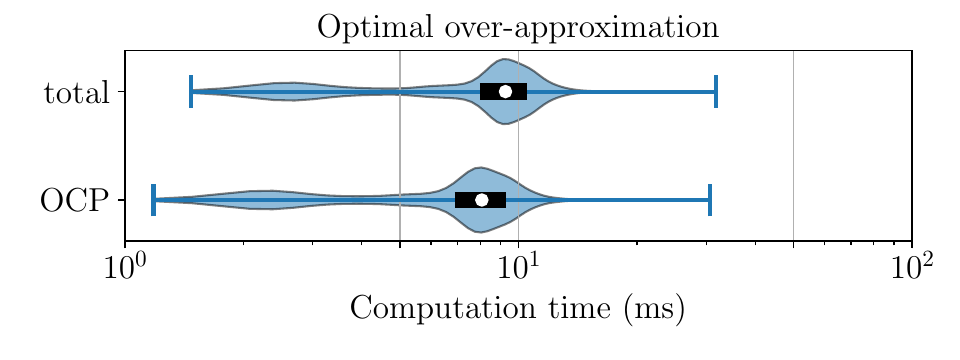}
		\label{fig:expo-narrow-real-robot-timings}} \\
    \vspace{-10pt}
    \subfloat
    {\includegraphics[width=0.98\linewidth]
        {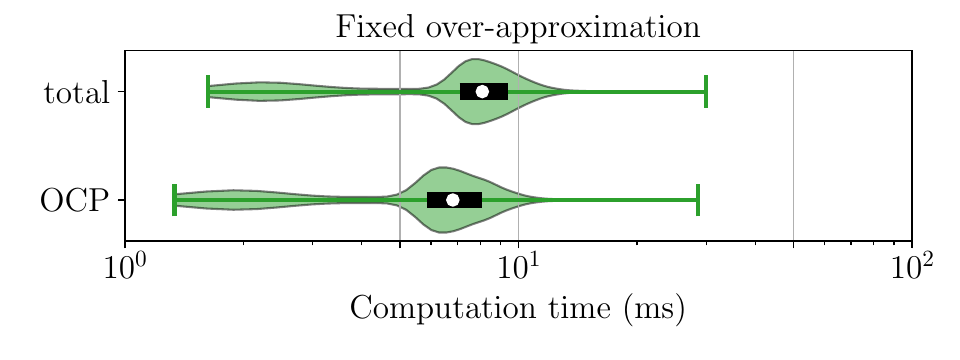}
        \label{fig:zo-narrow-real-robot-timings}}
    \caption{Computation time in real-world experiments (maximum two QP iterations).
    The white circle is the median.
    The black bar goes from the lower to the upper quartile.}
	\label{fig:narrow-real-results-timings}
    \end{minipage}
    \vspace{-5pt}
\end{figure*}

\subsubsection*{MPC Simulation Results}
Figure~\ref{fig:sim-results} displays the simulated robot trajectory when we optimize the over-approximations within the OCP and iteratively solve QP subproblems until convergence.
The robot safely navigates through narrow passages (see Fig.~\ref{fig:zoom-in-sim-trajectory}) and reaches the goal position (see Fig.~\ref{fig:sim-environment}).
It is noteworthy that certain sections of the path feature exceptionally narrow passages.
At around eleven seconds in the simulation, the available free space on either side of the robot is only a few centimeters.

\subsubsection*{Suboptimality}
We evaluate the suboptimality resulting from fixing the parameterization of the over-approximations in OCPs as discussed in Sec.~\ref{sec:practical-considerations-fixed-gamma}.
While running the MPC simulation with the over-approximations being optimized,
    we solve another OCP with the fixed over-approximations.
The relative additional cost is the increase in the optimal objective due to the fixed over-approximations divided by the optimal objective value for the optimal over-approximations.

We also evaluate the suboptimality induced by the fixed separating hyperplanes, which also retain collision-freeness.
To this end, let ${\eta\in\bbR^{n_{\mathrm{w}}}}$ parameterize the separating hyperplane.
The collision-avoidance constraint is formulated by $\eta^\top \left(p_{\mathrm{c}}(x_k) \! \shortminus \! t_m\right) \shortminus \sqrt{\eta^\top M_m \eta} \shortminus \sqrt{\eta^\top \tilde{G}(x_k) \eta} > 0$ and $\eta^\top \eta \leq 1$.
We determine the fixed hyperplane $\hat{\eta}$ by finding the two points, one in each ellipsoid, the vector between which describes the shortest vector between the two ellipsoids.
The hyperplane perpendicular to this vector is chosen as the fixed hyperplane.
The suboptimality induced by fixing the separating hyperplanes is notably greater than the suboptimality due to fixing the over-approximation parameters, as can be seen in Fig.~\ref{fig:fixed-gamma-eta-suboptimality}.
For fixed separating hyperplanes, the median and the worst-case increase in the cost is 1.36\% and 335\% respectively.
In contrast, for fixed over-approximations, where the $\hat{\gamma}$ values are computed using~\eqref{eq:compute-fixed-gamma},
    the resulting median and worst-case increase is 0.11\% and 9.2\% respectively.

Besides the cost comparison, we evaluate the closed-loop MPC in simulation.
Determining the value of the over-approximation parameter $\hat{\gamma}$ using~\eqref{eq:compute-fixed-gamma} enables the robot to safely travel through the narrow passages and reach the target position.
The minimum distances to the four obstacles are summarized in Table~\ref{table:simulation-min-dist}.
\begin{table}[t]
	\vspace{11pt}
    \caption{Minimum distance to obstacles in MPC simulation. The parameters for the over-approximations are updated outside the OCPs using~\eqref{eq:compute-fixed-gamma} and fixed in the OCPs.}
    \label{table:simulation-min-dist}
    \vspace{-7pt}
    \begin{center}
    \begin{tabular}{@{}lrrrr@{}}
    \hline
    \hline
    Obstacle Index & $m=1$ & $m=2$ & $m=3$ & $m=4$ \\
    \hline
    Distance\,($\mu$m) & 2007.5 & 171.2 & 67.3 & 1574.9 \\
    \hline
    \hline
    \end{tabular}
    \end{center}
    \vspace{-5pt}
\end{table}

\begin{table}[t]
    \caption{Computation time for solving OCPs in MPC simulation.
    The time for computing the value of $\hat{\gamma}$ and $\hat{\eta}$ is excluded.
    The maximum number of SQP iterations is two.}
    \label{table:simulation-cmpt-time}
    \vspace{-7pt}
    \begin{center}
    \begin{tabular}{@{}lrrrr@{}}
    \hline
    \hline
    Time\,(ms) &optimal $\gamma$ & fixed $\hat{\gamma}$ & optimal $\eta$ & fixed $\hat{\eta}$ \\
    \hline
    median & 1.53 & 1.24 & 1.74 & 1.16\\
    90\%   & 1.72 & 1.44 & 2.22 & 1.32\\
    worst  & 2.01 & 1.73 & 2.57 & 1.50\\
    \hline
    \hline
    \end{tabular}
\end{center}
\vspace{-25pt}
\end{table}

\subsubsection*{Computation Time}
We assess the solution time for the OCPs including the Minkowski-sum-based constraint formulation
    together with a comparison to the separating-hyperplane approach.
When the maximum number of iterations is sufficiently large to allow the SQP to fully converge,
    the Minkowski-sum-based formulation overall performs better than the separating-hyperplane approach (see Fig.~\ref{fig:computation-time}).
Fixing over-approximation parameters $\hat{\gamma}$ and fixing the separating hyperplanes $\hat{\eta}$ allow a significant decrease in the computation time.
The median computation times of the optimal over-approximations and the fixed over-approximations are 81.9\,ms and 5.6\,ms respectively.

It is not uncommon that the optimization process is terminated in early QP iterations for real-time capability~\cite{Diehl2005}.
Here we evaluate the computation time when solving the OCPs for maximum two iterations.
The results are reported in Table~\ref{table:simulation-cmpt-time}.
Fixing the over-approximations results in a decrease in the computation time by approximately 19\%.
While early termination greatly enhances real-time feasibility,
    the solutions are often not optimal and occasionally violate the constraints.
In consequence, safety margins need to be incorporated to ensure collision-free motions in practice.

\begin{figure}
    \centering
    \includegraphics[width=0.85\linewidth]
		{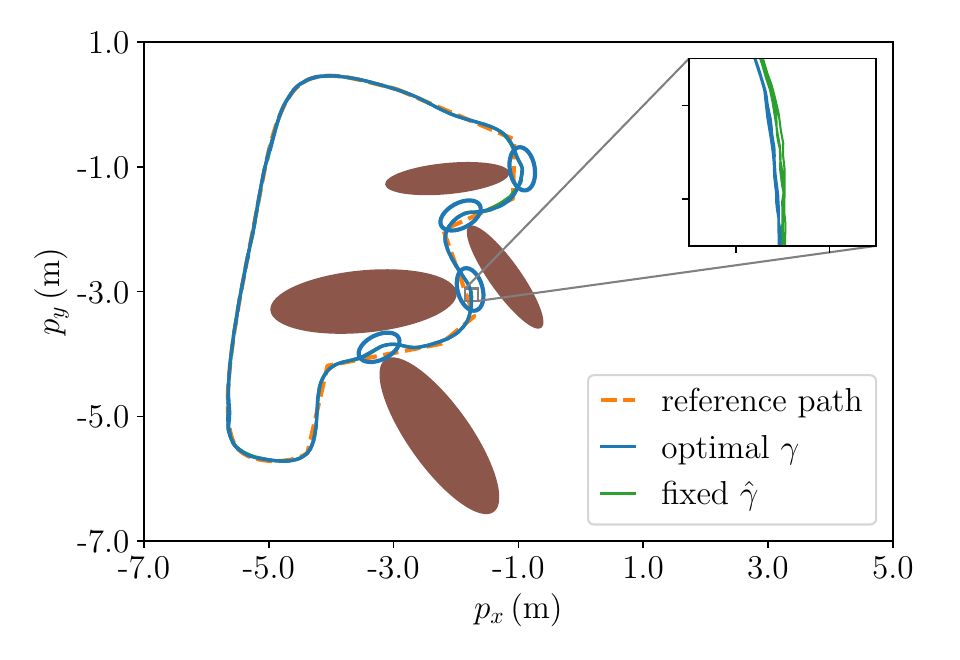}
	\vspace{-8pt}
    \caption{Robot trajectory in real-world experiments.
    The blue solid-line ellipsoids depict the differential-drive robot at different time instants.
    The brown ellipsoids are the obstacles.}
    \label{fig:real-robot-trajectory}
    \vspace{-15pt}
\end{figure}

\subsection{Real-World Experiments}
\subsubsection*{Experiment Setting}

The MPC is implemented as a Nav2 controller plugin written in C++~\cite{Macenski2020}.
The controller operates at a frequency of 20\,Hz.
The maximum number of SQP iterations is set to two.
As in simulation,
    the reference trajectory is provided by a time-optimal path tracking module.
Four virtual ellipsoidal obstacles are positioned in proximity to the path (see Fig.~\ref{fig:real-robot-trajectory}).
The obstacles are slightly more separated than in simulation.
The MPC has the knowledge of the ground-truth location of the obstacles while the path-tracking module that generates the reference trajectory is unaware of the obstacles.
The robot dynamics, in particular the robot motor controller, is not modeled in the~\eqref{eq:robot-model}, resulting in plant-model mismatch.
A safety margin of $0.01$ is added to the left-hand side of the collision-avoidance constraint~\eqref{eq:constr-formulation}.

\subsubsection*{Results}
The robot trajectory in the real-world experiments is plotted in Fig.~\ref{fig:real-robot-trajectory}.
For both the optimal and the fixed over-approximation approaches, the robot safely follows the reference path,
    adjusting its trajectory when potential collisions are imminent.
The two trajectories are barely distinguishable.
Only in some small sections along the path, the fixed over-approximation approach yields a slightly larger distance to the obstacle compared to the optimal case (see the zoomed-in region).
In the absence of collision risks,
    the robot adheres to the reference path.
The robot completes three cycles along the reference path.
The trajectories of different cycles overlay.

The computation time is reported in Fig.~\ref{fig:narrow-real-results-timings}.
The total computation time includes the durations for solving the OCP,
    generating the reference trajectory, and computing the value of $\hat{\gamma}$ for fixed over-approximation parameterization~\eqref{eq:compute-fixed-gamma}.
The median of the total computation time is 9.27\,ms for the optimal over-approximations and 8.10\,ms for the fixed case.
The computation manages to complete within the controller frequency, i.e., 50\,ms, for our test cases.

%% file: root.bbl
\begin{thebibliography}{10}

\bibitem{Brito2019}
B.~Brito, B.~Floor, L.~Ferranti, and J.~Alonso-Mora, ``Model predictive
  contouring control for collision avoidance in unstructured dynamic
  environments,'' {\em IEEE Robotics and Automation Letters}, vol.~4,
  pp.~4459--4466, Oct. 2019.

\bibitem{Zhang2021}
X.~Zhang, A.~Liniger, and F.~Borrelli, ``Optimization-based collision
  avoidance,'' {\em IEEE Trans. on Control Systems Technology}, vol.~29,
  pp.~972--983, May 2021.

\bibitem{Dietz2023}
C.~Dietz, S.~Albrecht, A.~Nurkanović, and M.~Diehl, ``Efficient collision
  modelling for numerical optimal control,'' in {\em Eur. Control Conf. (ECC)},
  IEEE, June 2023.

\bibitem{Reiter2024}
R.~Reiter, K.~Baumgärtner, R.~Quirynen, and M.~Diehl, ``Progressive smoothing
  for motion planning in real-time {NMPC},'' in {\em Eur. Control Conf. (ECC)},
  IEEE, June 2024.

\bibitem{Gao2023}
Y.~Gao, F.~Messerer, J.~Frey, N.~van Duijkeren, and M.~Diehl, ``Collision-free
  motion planning for mobile robots by zero-order robust optimization-based
  {MPC},'' in {\em Eur. Control Conf. ({ECC})}, {IEEE}, June 2023.

\bibitem{Birgin2015}
E.~G. Birgin, R.~D. Lobato, and J.~M. Martínez, ``Packing ellipsoids by
  nonlinear optimization,'' {\em J. of Global Optim.}, vol.~65, pp.~709--743,
  Dec. 2015.

\bibitem{Schneider2004}
P.~J. Schneider and D.~Eberly, {\em Geometric Tools for Computer Graphics}.
\newblock USA: Elsevier Science Inc., Sept. 2002.

\bibitem{Eberly2008}
D.~Eberly, ``Distance between ellipses in 2d,'' Mar. 2008.

\bibitem{Iwata2015}
S.~Iwata, Y.~Nakatsukasa, and A.~Takeda, ``Computing the signed distance
  between overlapping ellipsoids,'' {\em SIAM J. on Optim.}, vol.~25,
  pp.~2359--2384, Jan. 2015.

\bibitem{Alfano2003}
S.~Alfano and M.~L. Greer, ``Determining if two solid ellipsoids intersect,''
  {\em J. of Guidance, Control, and Dynamics}, vol.~26, pp.~106--110, Jan.
  2003.

\bibitem{Kallrath2013}
J.~Kallrath and S.~Rebennack, ``Cutting ellipses from area-minimizing
  rectangles,'' {\em J. of Global Optim.}, vol.~59, pp.~405--437, Dec. 2013.

\bibitem{Nocedal2006}
J.~Nocedal and S.~J. Wright, {\em Numerical optimization}.
\newblock Springer series in operations research and financial engineering, New
  York, NY: Springer, second~ed., 2006.

\bibitem{Rockafellar1970}
R.~T. Rockafellar, {\em Convex Analysis}.
\newblock Princeton University Press, Dec. 1970.

\bibitem{Houska2011}
B.~Houska, {\em Robust Optimization of Dynamic Systems}.
\newblock PhD thesis, KU Leuven, 2011.

\bibitem{Kurzhanskiy2007}
A.~A. Kurzhanskiy and P.~Varaiya, ``Ellipsoidal techniques for reachability
  analysis of discrete-time linear systems,'' {\em IEEE Trans. on Automatic
  Control}, vol.~52, pp.~26--38, Jan. 2007.

\bibitem{Verschueren2021}
R.~Verschueren, G.~Frison, D.~Kouzoupis, J.~Frey, N.~v. Duijkeren, A.~Zanelli,
  B.~Novoselnik, T.~Albin, R.~Quirynen, and M.~Diehl, ``acados—a modular
  open-source framework for fast embedded optimal control,'' {\em Math.
  Program. Computation}, vol.~14, pp.~147--183, Oct. 2021.

\bibitem{Frison2020}
G.~Frison and M.~Diehl, ``{HPIPM}: a high-performance quadratic programming
  framework for model predictive control,'' {\em {IFAC}-{PapersOnLine}},
  vol.~53, no.~2, pp.~6563--6569, 2020.

\bibitem{Daniel2010}
K.~Daniel, A.~Nash, S.~Koenig, and A.~Felner, ``Theta*: Any-angle path planning
  on grids,'' {\em J. of Artificial Intelligence Res.}, vol.~39, pp.~533--579,
  Oct. 2010.

\bibitem{Bobrow1985}
J.~E. Bobrow, S.~Dubowsky, and J.~S. Gibson, ``Time-optimal control of robotic
  manipulators along specified paths,'' {\em The Int. J. of Robotics Res.},
  vol.~4, no.~3, pp.~3--17, 1985.

\bibitem{Diehl2005}
M.~Diehl, H.~G. Bock, and J.~P. Schlöder, ``A real-time iteration scheme for
  nonlinear optimization in optimal feedback control,'' {\em SIAM J. on Control
  and Optim.}, vol.~43, pp.~1714--1736, Jan. 2005.

\bibitem{Macenski2020}
S.~Macenski, F.~Martin, R.~White, and J.~Ginés~Clavero, ``The marathon 2: A
  navigation system,'' in {\em IEEE/RSJ Int. Conf. on Intelligent Robots and
  Systems (IROS)}, Oct. 2020.

\end{thebibliography}
